# Vehicle Detection in Aerial Images

Michael Ying Yang[a], Wentong Liao[b], Xinbo Li[b], Bodo Rosenhahn[b]

[a]*ITC Faculty of Geo-Information Science and Earth Observation, University of Twente*
[b]*Institute of Information Processing, Leibniz University Hannover*

**Abstract**

The detection of vehicles in aerial images is widely applied in many applications. Comparing with object detection in the ground view images, vehicle detection in aerial images remains a challenging problem because of small vehicle size, monotone appearance and the complex background. In this paper, we propose a novel double focal loss convolutional neural network framework (DFL-CNN). In the proposed framework, the skip connection is used in the CNN structure to enhance the feature learning. Also, the focal loss function is used to substitute for conventional cross entropy loss function in both of the region proposed network and the final classifier. We further introduce the first large-scale vehicle detection dataset ITCVD with ground truth annotations for all the vehicles in the scene. We demonstrate the performance of our model on the existing benchmark DLR 3K dataset as well as the ITCVD dataset. The experimental results show that our DFL-CNN outperforms the baselines on vehicle detection.

## 1. Introduction

The detection of vehicles in aerial images is widely applied in many applications, *e.g.* traffic monitoring, vehicle tracking for security purpose, parking lot analysis and planning, *etc*. Therefore, this topic has caught increasing attention in both academic and industrial fields (Gleason et al., 2011; Liu and Mattyus, 2015;

*Corresponding author. Tel.: +31 53 489 2916; fax: +31 53 487 4335.
*Email addresses:* michael.yang@utwente.nl (Michael Ying Yang),
liao@tnt.uni-hannover.de (Wentong Liao), xinbo@tnt.uni-hannover.de (Xinbo Li),
rosenhahn@tnt.uni-hannover.de (Bodo Rosenhahn)

Chen et al., 2016). However, compared with object detection in ground view images, vehicle detection in aerial images has a lot of different challenges, such as much smaller scale, complex backgrounds and the monotonic appearance. See Figure 1 for an illustration.

Before the emergence of deep learning, hand-crafted features combined with a classifier are the mostly adopted ideas to detect vehicles in aerial images (Zhao and Nevatia, 2003; Gleason et al., 2011; Liu and Mattyus, 2015). However, the hand-crafted features lack generalization ability, and the adopted classifiers need to be modified to adapt the of the features. Some previous works also attempted to use shallow neural network (LeCun et al., 1990) to learn the features specifically for vehicle detection in aerial images (Cheng et al., 2012; Chen et al., 2014). However, the representational power of the extracted features are insufficient and the performance meets the bottleneck. Furthermore, all of these methods localize vehicle candidates by sliding window search. It's low efficient and leads to costly and redundant computation. The window's sizes and sliding steps must be carefully chosen to adapt the varieties of objects of interest in dataset.

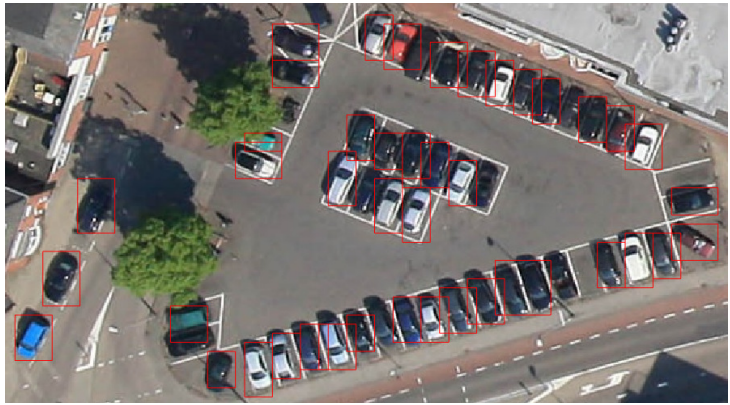

Figure 1: Vehicles detection results on the proposed dataset.

In recent years, deep convolutional neural network (DCNN) has achieved great successes in different tasks, especially for object detection and classification classification (Krizhevsky et al., 2012; LeCun et al., 2015). In particular, the series of methods based on region convolutional neural network (R-CNN) (Girshick et al., 2014; Girshick, 2015; Ren et al., 2015) push forward the progress of object detection significantly. Especially, Faster-RCNN (Ren et al., 2015) proposes the region proposal network (RPN) to localize possible object instead of traditional sliding window search methods and achieves the state-of-the-art performance in different datasets in terms of accuracy. However, these existing state-of-the-art detectors cannot be directly applied to detect vehicles in aerial images, due to the different characteristics of ground view images and aerial view images (Xia et al., 2017). The appearance of the vehicles are monotone, as shown in Figure 1. It's difficult to learn and extract representative features to distinguish them from other objects. Particularly, in the dense park lot, it is hard to separate individual vehicles. Moreover, the background in the aerial images are much more complex than the nature scene images. For examples, the windows on the facades or the special structures on the roof, these background objects confuse the detectors and classifiers.Furthermore, compared to the vehicle sizes in ground view images, the vehicles in the aerial images are much smaller (ca. $50 \times 50$ pixels) while the images have very high resolution (normally larger than $5000 \times 2000$ pixels). Lastly, large-scale and well annotated dataset is required to train a well performed DCNN methods. However, there is no public large-scale dataset such as ImageNet (Deng et al., 2009) or ActivityNet (Caba Heilbron et al., 2015), for vehicle detection in aerial images.

To address these problems, we propose a specific framework for vehicle detection in aerial images, as shown in Figure 2. The novel framework is called double focal loss convolutional neural network (DFL-CNN), which consists of three main parts: 1) A skip-connection from the low layer to the high layer is added to learn features which contains rich detail information. 2) Focal loss function (Lin et al., 2017) is adopted in the RPN instead of traditional cross entropy. This modification aims at the class imbalance problem when RPN determine whether a proposal is likely an object of interest. 3) Focal loss function replaces the cross entropy in the classifier. It's used to handle the problem of easy positive examples and hard negative examples during training. Furthermore, we introduce a novel large-scale and well annotated dataset for quantitative vehicle detection evaluation - ITCVD. Towards this goal, we collected 173 images with 29088 vehicles, where each vehicle in the ITCVD dataset is manually annotated using a bounding box. The performance of the proposed method is demonstrated with respect to a representative set of state-of-the-art baselines, leveraging the proposed ITCVD dataset and DLR 3K dataset (Liu and Mattyus, 2015). We make our code and dataset online available.

## 2. Related Work

Object detection and classification have been the central topics in the computer vision and photogrammetry literature. Most of the existing methods can be roughly divided into three main steps: candidate region proposal, feature extraction and classification.

To generate the regions which likely contains the object of interest, many methods employ sliding-window search strategy (Felzenszwalb et al., 2010; Liu and Mattyus, 2015; Chen et al., 2016). These methods used windows with varied scales and ration to scan through the image with fixed step size. Sliding-window search strategy has high computation and time complexity, and most of the windows are redundant. Uijlings et al. (2013) proposed the algorithm dubbed *Selective Search* to generate possible object locations. This method combines the merits of both an exhaustive search and segmentation. It's widely adopted to combine with DCNN methods for object detection, such as Girshick et al. (2014); Girshick (2015). Ren et al. (2015) proposed the *region proposal network* (RPN) and then became the most popular method for region proposal.

Before classification, features are extracted within each region candidate. Kembhavi et al. (2011) used SIFT features for vehicle detection. Gleason et al. (2011); Han et al. (2006) employed HoG features, while Bai et al. (2006) adopted Haar-like features for this task. Even though their methods reported good results, such hand-crafted features are insufficient to separate vehicles from the complex background. Recently, DCNN based methods have achieved great successes in object detection and classification (Krizhevsky et al., 2012; Girshick et al., 2014; Tang et al., 2017; Carlet and Abayowa, 2017) .

Finally, the extracted features are feed to a classifier. Support Vector Machine (SVM) and Random Forest (RF) are two of the most popular classifiers (Zhao and Nevatia, 2003; Gleason et al., 2011; Liu and Mattyus, 2015; Rey et al., 2017) because of their high efficiency and robustness. Until now, they are also employed as the final classifier of some CNN based methods (Girshick et al., 2014). Recently, softmax is the first choice for the classifier of DCNN based methods because it provides normalized probabilistic prediction. Then the cross entropy (CE) is used to calculate the loss for propagation to update the parameters of the network (LeCun et al., 2015).

The methods which consist of these three steps are well known as *two-stage* methods: candidate region proposal at the first stage and object classification at the second stage. The CNN based two-stage methods achieve the state-of-the-art performance in terms of accuracy. In contrast, the methods which do not need an additional operation for region proposals, such as YOLO (Redmon et al., 2016) and SSD (Liu et al., 2016), are one-stage-methods. They perform much faster than two-stage methods with compromise of accuracy. Especially, their performance of detecting objects in small scale is very poor. This demerit limits their application for vehicle detection in aerial images. Therefore, we utilize two-stage method in our framework.

For training a well performing CNN based methods which has millions of parameters, large dataset is the key factor. In the past, some well known large-scale datasets for different tasks are published, *e.g.*ImageNet (Deng et al., 2009) for object classification, Cityscapes dataset (Cordts et al., 2016) for semantic segmentation, *etc*. All of them consists of tens of thousands images for training the model. Even though many existing benchmark datasets contain varieties of vehicles, they are collected in the ground view. These datasets are not applicable to train a framework for vehicle detection in aerial images. There are also existing some well annotated dataset for aerial images, such as the VEDAI dataset (Razakarivony and Jurie, 2016) and DLR 3K dataset (Liu and Mattyus, 2015). However, the objects in the VEDAI dataset are relative easy to detect because of the small number of vehicles which sparsely distribute in the images, and the background is simple. The more challenging and realistic DLR 3K dataset contains totally 20 aerial images with resolution of $5616 \times 3744$. 10 images (3505 vehicles) are used for training. Such number of training samples seems too small for training a CNN model. In comparison with aforementioned datasets, our new dataset ITCVD provides 135 images with 23543 vehicles for training the network.

## 3. Proposed Framework

An overview of the proposed framework is illustrated in Figure 2. It's modified based on the standard Faster R-CNN (Ren et al., 2015). We refer readers to Ren et al. (2015) for the general procedure of object detection. In this work, we choose ResNet (He et al., 2016) as the backbone structure for feature learning, because of its high efficiency, robustness and effectiveness during training (Canziani et al., 2016).

### 3.1. Skip Connection

It has been proven in the task of semantic segmentation that, features from the shallower layers retain more detail information (Long et al., 2015). In the task of object detection, the sizes of vehicles in aerial images are ca. $30 \times 50$ pixels, assuming 10cm GSD. The size of the output feature maps of the ResNet from the 5*th* pooling layers is only one 32nd of the input size (He et al., 2016). The shorter edges of most vehicles are very small when they are projected on the feature maps after the 5*th* pooling layer. So, they will be ignored because their sizes are rounded up. Furthermore, pooling operation leads to significant loss of detailed information. For densely parked area, it is difficult to separate individual vehicles. For example, as shown in Figure 3, the extracted features from the shallow layer (Figure 3b) have richer detailed information than the features from the deeper layer (Figure 3c). In the case of densely parked area (Figure 3a), the detail information play an important to separate the individual vehicles from each other. Therefore, we fuse the features from the shallow layers, which contain more detail information, with the features learned by deeper layers, which have more representative abilities, to precisely localize detected individual vehicle. This skip-connected CNN architecture is illustrated in Figure 4. The image fed to the network is $752 \times 674$ pixels. The size of the feature maps from the $4th$ and $5th$ pooling layers are $42 \times 47 \times 1024$ and $21 \times 24 \times 2048$ respectively. To fuse them together, the smaller feature maps are upsampled to the size of $42 \times 47 \times 2048$, and then reduced the feature channels into 1024 by a $1 \times 1$ convolution layer. Then the two feature maps are concatenated as the skip-connected feature maps.

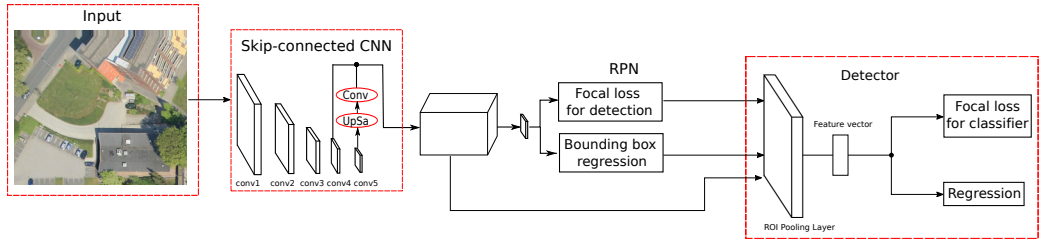

Figure 2: The overview of the proposed framework DFL-CNN. It consists of three main parts: 1) A skip-connection from the low layer to the high layer is added to learn features which contains rich detail information. 2) Focal loss function (Lin et al., 2017) is adopted in the RPN instead of traditional cross entropy. 3) Focal loss function replaces the cross entropy in the classifier.

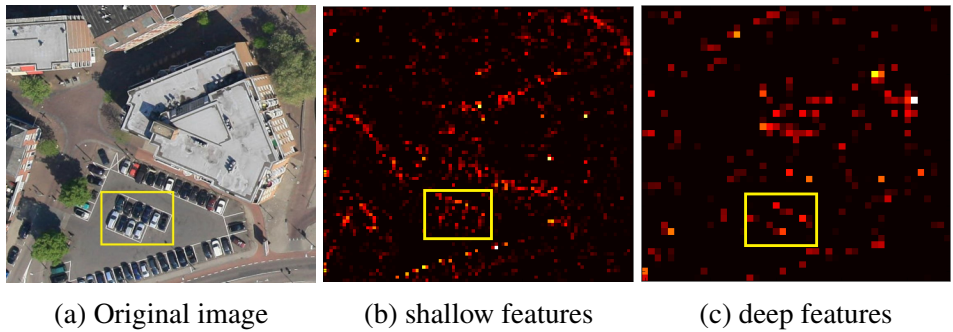

(a) Original image     (b) shallow features     (c) deep features

Figure 3: Comparison of the extracted features from the 4th pooling layer (b) and the 5th pooling layer (c). They are illustrated in heat map. The yellow bounding box indicate the corresponding region in the original image and the feature maps.

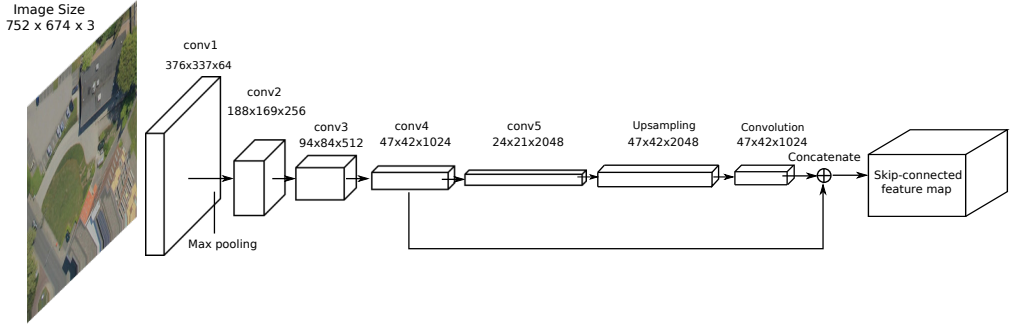

Figure 4: Structure of skip-connected CNN. The feature maps from the conv5 are upsampled to the same size as the feature maps from conv4. Then, the number of the feature channels are reduced by $1 \times 1$ convolution layer into 1024. Finally, the feature maps from conv4 and conv5 are concatenated.

*3.2. Loss function*

*Cross entropy* (CE) is the most popular loss function used for object classification. It can reduce the imbalance between positive and negative samples. But it is not good enough to train classifier for distinguishing easy and hard classified examples. This problem becomes more significant in the task of vehicle detection in the aerial images because of the monotone appearance of target objects and the complex background. For example, windows on the facade may have very similar appearance as the cars.

*Focal loss function* is original proposed by Lin et al. (2017) to dedicate the class imbalance problems for the one-stage object detectors, such as YOLO (Redmon et al., 2016) and SSD (Liu et al., 2016). As discussed in the paper, a one-stage detector suffers from the extreme foreground-background class imbalance because of the dense candidates which cover spatial positions, scales, and aspect ratios. A two-stage detector handles this challenge in the first stage: candidates proposal, *e.g.* RPN (Ren et al., 2015), most of the candidates which are likely to be the background are canceled, and then the second stage: classifier works on much sparser candidates. However, in the scenes with dense objects of interest, *e.g.*, the parking cars in Figure 1, even the state-of-the-art candidates proposal method RPN is not good enough to filter the dense proposals in two aspects: 1) many of the dense proposals cover two vehicles and have high overlap with the ground truth, which makes it hard for the proposal methods to determine whether they are background objects. 2) Too many background objects interfere the training. It is hard to select the negative samples which are very similar as the vehicles to enhance the detector/classifier to distinguish them from the positive samples. Inspired by the idea in Lin et al. (2017), we proposed to use the *focal loss function* instead of the conventional CE loss both in the region proposal and the classification stages, dubbed as *double focal loss-CNN* (DFL-CNN). For better understanding, let's have a brief review on *focal loss function*.

The traditional CE loss for classification (for convenient discussion, we take the binary classification as example) is formally defined as:

$$L_{CE}(p,y) = -\log(p_t), \qquad (1)$$

with $p_t = \begin{cases} p & \text{if } y=1 \\ 1-p & \text{otherwise,} \end{cases}$ where $p$ is the predicted probability of given candidate having label $+1$ and $y$ is its ground truth label and $y \in \{-1,+1\}$.

When add a modulating factor $(1-p_t)^\gamma$ with tunable focusing parameter $\gamma \geq 0$ to the CE loss, the loss function becomes the so call *focal loss* (FL):

$$L_{FL}(p_t) = -(1-p_t)^\gamma \log(p_t) \qquad (2)$$

The focal loss has two main properties: 1) The loss is unaffected by misclassified examples which have small $p_t$ when the modulating factor is near 1. In contrast, when $p_t \to 1$, the modulating factor is near 0, which down-weights the loss for well-classified examples. 2)When the focusing parameter $\gamma$ is increased, the effect of modulating factor is also increased. CE is the special case of $\gamma = 0$. Intuitively, the contribution of the easy examples are reduced while the ones from hard examples are enhanced during the training. For example, with $\gamma = 2$ [1], the focal loss of an example classified with $p_t = 0.9$ is 1% of the CE loss and 0.1% of it when $p_t = 0.968$. If an example is misclassified ($p_t < 0.5$), its importance for training is increased by scaling down its loss 4 times.

*3.3. Double Focal Loss CNN*

In our DFL-CNN framework, we add a skip connection to fuse the features from the lower (conv4) and higher (conv5) layers, and adopt focal loss function both in the RPN layer and the final classification layer to overcome the class imbalance and the easy/hard examples challenges in our task.

---

[1]$\gamma$ is set to 2 in our experiments.

As discussed in Section 3.1, the final feature maps are 1/16 of the original images. Therefore, each pixel in the feature maps corresponds an region of $16 \times 16$ pixels. To generate candidates proposal, centered on each pixel in the feature maps, 9 anchors in 3 different areas ($30^2$, $50^2$, $70^2$) and 3 different ratios (1:1, 2:1 and 1:2) are generated on the original input image. Every anchor is labeled as either positive or negative sample based on the Intersection-over-Union (IoU) with ground truth. The IoU is formally defined as: $\text{IoU} = \frac{area(\text{Proposal} \cap \text{Ground Truth})}{area(\text{Proposal} \cup \text{Ground Truth})}$, where the numerator is the overlapping area of box of candidate and the ground truth box, and the denominator represents the union of them. The proposals which have the IoU more than 0.7, are labeled as positive samples and the ones whose IoU are smaller than 0.1 are labeled as the negative samples. Other proposals are discarded. All the proposals exceeding the boundary of the image are also discarded. During training, each mini-batch consists of 64 positive samples and 64 negative samples.

The loss function for training the RPN using focal loss is defined as:

$$L_{RPN}(\{p_i\},\{t_i\}) = \frac{1}{N_{cls}} \sum_i L_{cls-FL}(p_i, p_i^*) + \lambda \frac{1}{N_{reg}} \sum_i p_i^* L_{reg}(t_i, t_i^*) \qquad (3)$$

where $L_{cls-FL}$ is the focal loss for classification, as defined in Equation (2) and $L_{reg}$ is the loss for bounding box regression. $p_i$ is the predicted probability of proposal $i$ belonging to the foreground and $p_i^*$ is its ground truth label. $N_{cls}$ denotes the total number of samples and $N_{reg}$ is the total number of positive samples. $\lambda$ is used to weight the loss for bounding box regression [2]. The smooth $L_1$ loss function is adopted for $L_{reg}$ as in Ren et al. (2015):

$$L_{reg}(t_i, t_i^*) = f_{smooth}(t_i - t_i^*), \qquad (4)$$

with $f_{smooth}(j) = \begin{cases} 0.5 j^2 & \text{if } |j| < 1 \\ |j| - 0.5 & \text{otherwise.} \end{cases}$

$t = (t_x, t_y, t_w, t_h)$ is the normalized information of the bounding boxes of the positive sample and $t^*$ is its ground truth. Each of the entries is formally defined

---

[2]$\lambda$ is set to 15 in our experiments. Because the size of final feature maps is $47 \times 42$ and totally 128 anchors are chosen, therefore the ratio is ca. 15.

as:

$$\begin{aligned}
t_x &= (P_x - A_x)/A_w, & t_y &= (P_y - A_y)/A_h, \\
t_w &= \log(P_w/A_w), & t_h &= \log(P_h/A_h), \\
t_x^* &= (P_x^* - A_x)/A_w, & t_y^* &= (P_y^* - A_y)/A_h, \\
t_w^* &= \log(P_w^*/A_w), & t_h^* &= \log(P_h^*/A_h),
\end{aligned} \quad (5)$$

where $(P_x, P_y)$ is the center coordinate of the predicted bounding box and $(P_w, P_h)$ is its predicted width and height, and so as the the bounding box information of the anchors $A = (A_x, A_y, A_w, A_h)$. $P^*$ is the ground truth bounding box information.

The RPN layer output a set of candidates which are likely to be the objects of interest, *i.e.* vehicles in this work, and there predicted bounding boxes. Then, the features covered by these bounding boxes are cropped out from the feature maps and go through the region of interest (ROI) pooling layer to get a fix the size of features.

Finally, the final classifier subnet are fed with these features and classify their labels, and predict their bounding boxes further. The loss function of the classifier subnet for each candidate is formally defined as:

$$L_{classifier}(P,T) = L_{cls-FL}(P,P^*) + \lambda_2 P^* L_{reg}(T,T^*) \quad (6)$$

where $T$ is defined as:

$$\begin{aligned}
T_x &= (P_x - A_x)/A_w, & T_y &= (P_y - A_y)/A_h, \\
T_w &= \log(P_w/A_w), & T_h &= \log(P_h/A_h), \\
T_x^* &= (P_x^* - A_x)/A_w, & T_y^* &= (P_y^* - A_y)/A_h, \\
T_w^* &= \log(P_w^*/A_w), & T_h^* &= \log(P_h^*/A_h),
\end{aligned} \quad (7)$$

The $P_x$, $A_x$ and $P_x^*$ denote the bounding boxes of prediction results, anchors and ground truth. The other subscripts of $y$, $w$ and $h$ are the same as $x$. We set $\lambda_2 = 1$ to equal the influence of classification and bounding box prediction. During training, the classifier subnet is trained using positive and negative samples in ratio of 1 : 3, same as the conventional training strategy (Ren et al., 2015).

## 4. ITCVD Dataset

In this section, we introduce the new large-scale, well annotated and challenging ITCVD dataset. The images were taken from an airplane platform which flied over Enschede, The Netherlands, in the height of ca 330m above the ground (Slagboom en Peeters, 2017). The images are taken in both nadir view and oblique view, as shown in Figure 5. The tilt angle of oblique view is 45 degrees. The Ground Sampling Distance (GSD) of the nadir images is 10cm.

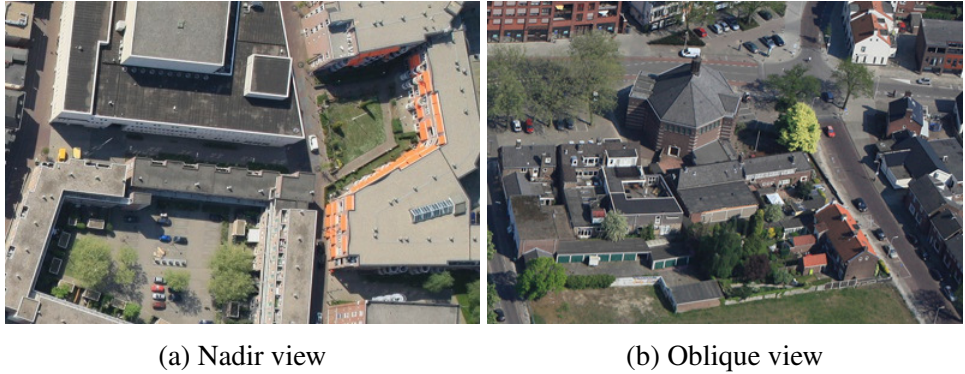

(a) Nadir view    (b) Oblique view

Figure 5: Example images in ITCVD dataset, which are taken in both nadir view (a) and oblique view (b).

The raw dataset contains 228 aerial images with high resolution of $5616 \times 3744$ pixels in JPG format. Because the images are taken consecutively with a small time interval, there is ca. 60% overlap between consecutive images. It is important to make sure that, the images used for training do not have common regions with the images that are used for testing. After careful manual selection and verification, 173 images are remained among which 135 images with 23543 vehicles are used for training and the remaining 38 images with 5545 vehicles for testing. Each vehicle in the dataset is manually annotated using a bounding box which is denoted as $(x, y, w, h)$, where $(x, y)$ is the coordinate of the left-up corner of the box, and $(w, h)$ is the width and height of the box respectively.

## 5. Experiments

In this section, we discuss about the experimental settings and datasets, in which we evaluate the proposed method and compare with the state-of-the-art object detectors.

*5.1. Dataset and experimental settings*

We evaluate our method in our ITCVD and DLR 3K datasets (Liu and Mattyus, 2015). The statistic information of the two datasets are listed in Table 1. The state-of-the-art object detector Faster R-CNN (Ren et al., 2015) is implemented in these datasets to provide a strong baseline.

|  | **Training set** | **Testing set** | **Image size** |
|---|---|---|---|
| ITCVD | 135 images (23543 vehicles) | 38 images (5545 vehicles) | 5616×3744 |
| DLR 3K | 10 images (3505 vehicles) | 10 images (5928 vehicles) | 5616×3744 |

Table 1: Statistic of ITCVD dataset and DLR 3K dataset (Liu and Mattyus, 2015).

TO save the GPU memory, each original image in the datasets are cropped into small patches uniformly. The resulting new image patches are in the size of $674 \times 752$ pixels. The coordinate information of annotation is also updated in the new cropped patches. In the DLR 3K dataset, each vehicle is annotated with a tightly fit bounding box. To adapt our experiment settings, the original annotation is transformed to a normal square bounding box which is expressed with its center point, height and width.

The deep learning models are implemented in Keras with TensorFlow Abadi et al. (2016) backend. The ResNet-50 network He et al. (2016) is used as the backbone CNN structure for feature learning for Faster R-CNN and our model. We use a learning rate of 0.00001 to train the RPN. Note that, other CNN structures, *e.g.*VGGnet Simonyan and Zisserman (2014) and Google Inception Szegedy et al. (2016), are also applicable in our framework. The CNN structure are pre-trained on ImageNet dataset Deng et al. (2009).

To evaluate the experimental results, the metrics of recall/precision rate and $F1$-score are used, which are formally defined as:

$$\text{Recall Rate (RR)} = \frac{\text{TP}}{\text{TP+FN}}, \tag{8}$$

$$\text{Precision Rate (PR)} = \frac{\text{TP}}{\text{TP+FP}}, \tag{9}$$

$$\text{F1-score} = \frac{2 \times \text{RR} \times \text{PR}}{\text{RR+PR}}. \tag{10}$$

where, *TP*, *FN*, *FP* denote the *true positive*, *false negative* and *false positive* respectively. Furthermore, the relationships between the *IoU* and *RR*, *PR* are also evaluated respectively.

*5.2. Results on ITCVD dataset*

We evaluated our method DFL-CNN in our challenging ITCVD dataset. The state-of-the-art object detector Faster R-CNN (Ren et al., 2015) is implemented to provide a strong baseline. In addition, traditional HOG + SVM method Dalal and Triggs (2005) is provided as a weak baseline.

Figure 6 depicts the relationship between recall rate and the precision rate of DFL-CNN, Faster R-CNN and HOG+SVM algorithms with different IoU in the ITCVD dataset. It is obvious that the CNN based methods (DFL-CNN in green curve and Faster R-CNN in red curve) are significantly better than the traditional method (HOG+SVM in black curve). In the relation between recall and precision, our DFL-CNN method also perform better than Faster R-CNN. According to these relationship curves, $IoU = 0.3$ is a good balance point for the following experimental settings, which reports high recall rate and precision at the same time. Note that, it is also a conventional setting in the task of object detection. The quantitative results of these three methods are given in Table 2 (the results are given with $IoU = 0.3$). We can see that, our method outperforms the others.

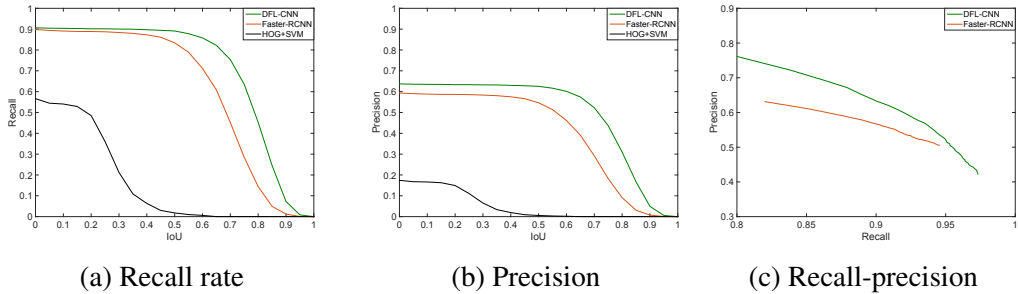

(a) Recall rate  (b) Precision  (c) Recall-precision

Figure 6: The relationship between IoU and recall rate (a), IoU and precision rate (b) and recall and precision (c) of DFL-CNN, Faster R-CNN, HOG+SVM in the ITCVD dataset respectively.

|  | **HOG+SVM** | **Faster R-CNN** | **DFL-CNN** |
|---|---|---|---|
| **Recall Rate** | 21.19% | 88.38% | **89.44%** |
| **Precision Rate** | 6.52% | 58.36% | **64.61%** |
| **F1-score** | 0.0997 | 0.7030 | **0.7502** |

Table 2: Comparison of baselines and the DFL-CNN method in ITCVD dataset.

To justify the gain by using skip connection and focal loss function, we conducted extensive experiments in ablation studies. First, we train two frameworks both using double focal loss function. But one of the framework has skip connection of the feature maps and the other one not. The qualitative results are shown in Figure 7. We can observe that, the bounding boxes predicted by the framework with skip connection of the feature maps are much more precise than those that predicted by the framework without skip connection. Individual vehicle is also separated better from others by using the shallow features. Then, we train two frameworks with the skip connection. But one of the framework is trained using CE as loss function and the other using double loss function. The qualitative results are shown in Figure 8. In the results given by CE-trained framework, many background objects that have similar appearances as vehicle are easily to be falsely detected as vehicle. The framework trained using double focal loss function distinguishes these hard negative samples much better from the real vehicles.

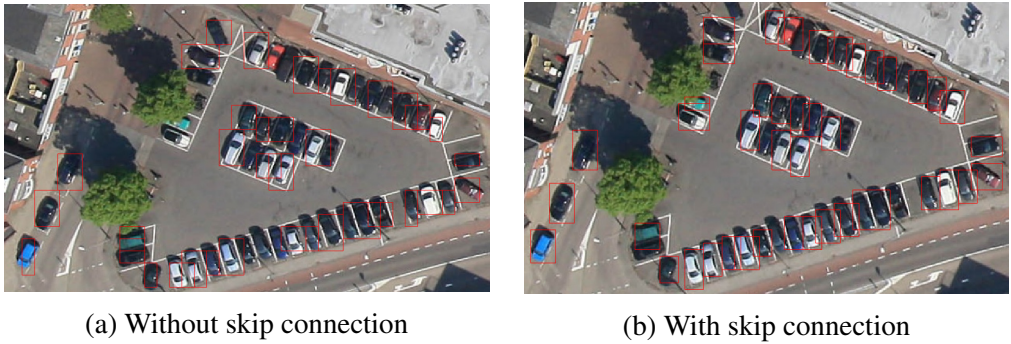

(a) Without skip connection  (b) With skip connection

Figure 7: Qualitative comparison of bounding box prediction of different frameworks that has no (a) and has (b) skip connection. Other settings are the same.

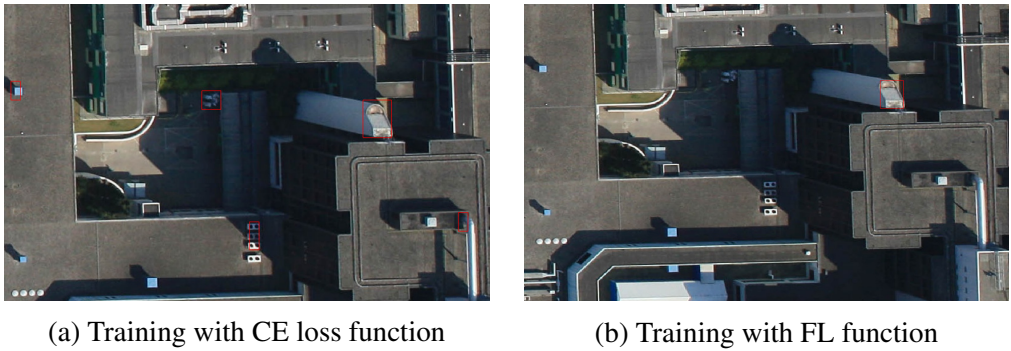

(a) Training with CE loss function  (b) Training with FL function

Figure 8: Qualitative comparison of vehicle detection of different frameworks that is trained using CE loss(a) and FL (b) function. Other settings are the same.

Figure 9 gives some examples of bad detection results of the proposed method. Even through our method achieves significant improvements in detection precision and recall rate than the baseline methods, our detector still misses to detect some obvious vehicles, especially in the crowded parking lot, as shown in Figure 9a. On the other hand, some objects which have very similar appearances as vehicle are also falsely detected, as shown in Figure 9b.

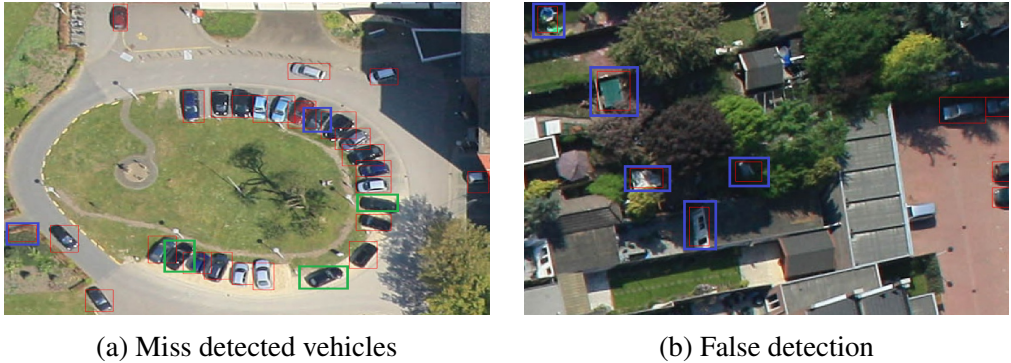

(a) Miss detected vehicles

(b) False detection

Figure 9: Qualitative examples of incorrect detection by our model. The boxes in red thin line denote the detection results, and the green boxes denote the missed detected vehicles while the blue boxes indicate the false positive prediction.

## 5.3. Results on DLR 3K dataset

We also evaluated our model in DLR 3K dataset (Liu and Mattyus, 2015). In Figure 10, the relationship between the recall rate and precision is depicted, both for the Faster-RCNN and the proposed method. Figure 10 also indicate that, our method outperform the standard Faster R-CNN in terms of recall rate and precision. In particular, we compared the performance of Faster R-CNN and DFL-

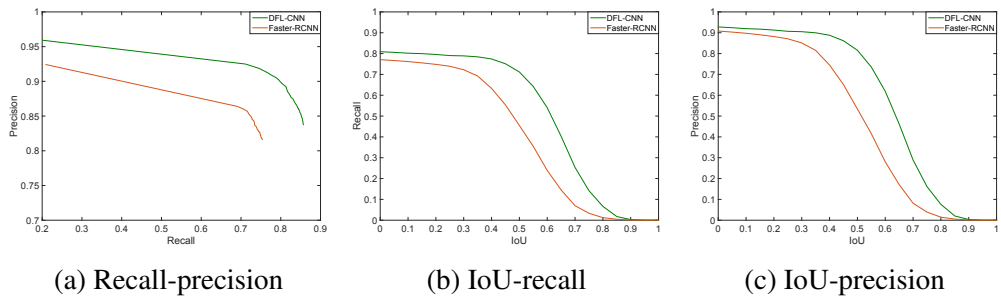

(a) Recall-precision

(b) IoU-recall

(c) IoU-precision

Figure 10: The relationship between recall and precision (a), IoU and precision (b) and IoU and recall rate (c) of DFL-CNN and Faster R-CNN in the DLR 3K dataset (Liu and Mattyus, 2015).

CNN in the case of densely parked vehicles in DLR 3K dataset, as shown in

Figure 11. From the qualitative results we can see that, DFL-CNN (Figure 11b) detected more individual vehicles and predicted more precise bounding boxes than Faster R-CNN (Figure 11a).

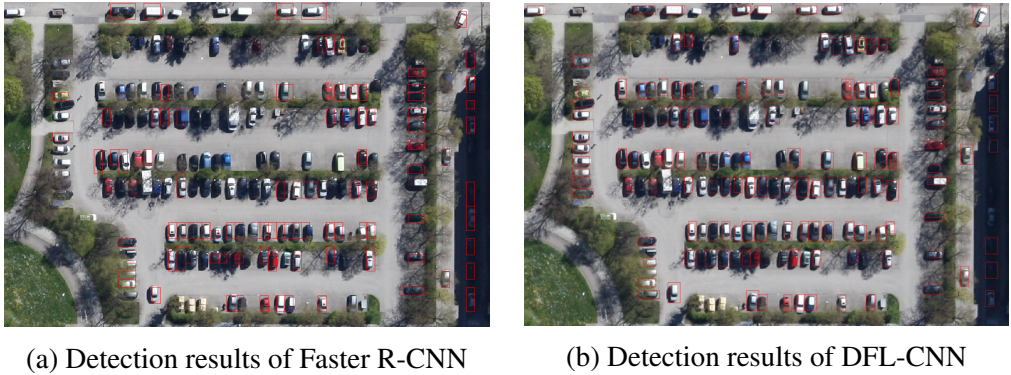

(a) Detection results of Faster R-CNN  (b) Detection results of DFL-CNN

Figure 11: Qualitative comparison of detection results given by Faster R-CNN (a) and DFL-CNN (b) in the DLR 3K dataset, respectively.

## 6. Conclusion

In this paper, we have proposed a specific framework DFL-CNN for vehicle detection in the aerial images. We fuse the features properties learned in the lower layer of the network (containing more spatial information) and the ones from higher layer (more representative information) to enhance the network's ability of distinguishing individual vehicles in a crowded scene. To address the challenges of class imbalance and easy/hard examples, we adopt focal loss function instead of the cross entropy in both of the region proposal stage and the classification stage. We have further introduced the first large-scale vehicle detection dataset ITCVD with ground truth annotations for all the vehicles in the scene. Compared to DLR 3K dataset, our benchmark provides much more object instances as well as novel challenges to the community. The experimental results show that our method outperforms the state-of-the-art in these two datasets. For future work, we will extend DFL-CNN to recognize the vehicle types and detect the vehicle orientations. We will also continue to update ITCVD dataset to grow in size and scope. We believe the new ITCVD dataset will promote the development of object detection algorithms in the photogrammetry community.

## Acknowledgments

The work is funded by DFG (German Research Foundation) YA 351/2-1 and RO 4804/2-1. The authors gratefully acknowledge the support. We thank Slagboom en Peeters for providing the aerial images.